\documentclass[letterpaper, 10 pt, conference]{ieeeconf}  

\IEEEoverridecommandlockouts                             
\usepackage{cite}
\usepackage{amsmath,amssymb,amsfonts}
\usepackage{algorithmic}
\usepackage{graphicx}
\usepackage{textcomp}
\usepackage{xcolor}
\usepackage{url}
\def\BibTeX{{\rm B\kern-.05em{\sc i\kern-.025em b}\kern-.08em
    T\kern-.1667em\lower.7ex\hbox{E}\kern-.125emX}}
    
\usepackage[linesnumbered,ruled]{algorithm2e}
\usepackage{xcolor}
\usepackage{paralist}
\newcommand{\norm}[1]{\left\lVert#1\right\rVert}

\let\labelindent\relax

\usepackage{amsthm}
\theoremstyle{plain}

\usepackage[linesnumbered,ruled]{algorithm2e}
\theoremstyle{remark}

\theoremstyle{definition}

\theoremstyle{plain}
\newtheorem{theorem}{Theorem}

\newtheorem{problem}{Problem}

\usepackage{dirtytalk}
\usepackage[caption=false]{subfig}
\usepackage{mathtools}

\title{\LARGE \bf
Fast Biconnectivity Restoration in Multi-Robot Systems for Robust Communication Maintenance
}

\author{Md. Ishat-E-Rabban$^{1}$, Guangyao Shi$^{2}$, and Pratap Tokekar$^{1}$
\thanks{Rabban and Shi have contributed equally.}
\thanks{\textsuperscript{$^{1}$}Department of Computer Science, University of Maryland, College Park, MD 20742, USA. email: ier@umd.edu, tokekar@umd.edu}
\thanks{\textsuperscript{$^{2}$}Department of Electrical and Computer Engineering, University of Southern California, Los Angeles, CA 90007, USA. email: shig@usc.edu.}
\thanks{This work is supported by the National Science Foundation under Grant No. 1943368, and the Office of Naval Research under Grant No. N000141812829.}
}

\begin{document}

\maketitle
\thispagestyle{empty}
\pagestyle{empty}

\begin{abstract}

Maintaining a robust communication network plays an important role in the success of a multi-robot team jointly performing an optimization task. A key characteristic of a robust multi-robot system is the ability to repair the communication topology itself in the case of robot failure. In this paper, we focus on the \textit{Fast Biconnectivity Restoration} (FBR) problem, which aims to repair a connected network to make it biconnected as fast as possible, where a biconnected network is a communication topology that cannot be disconnected by removing one node. We develop a Quadratically Constrained Program (QCP) formulation of the FBR problem, which provides a way to optimally solve the problem. We also propose an approximation algorithm for the FBR problem based on graph theory. By conducting empirical studies, we demonstrate that our proposed approximation algorithm performs close to the optimal while significantly outperforming the existing solutions. 


\end{abstract}


\section{Introduction}
\label{intro}

There is a growing trend in assigning a team of robots to perform coverage, exploration, patrolling, and similar cooperative optimization tasks. Compared to a single robot, a multi-robot system can extend the range of executable tasks, enhance task performance, and naturally provide redundancy for the system. Communication plays a key role in the successful deployment of a multi-robot system. To improve task performance, it is important to maintain communication among the robots by forming a connected network. However, maintaining a connected communication network is not adequate for multi-robot systems, because robots may fail, e.g., experience sensor or actuator malfunction, or undergo adversarial attack.

The existing works on ensuring robustness of multi-robot systems mostly focus on maintaining a $k$-connected network topology~\cite{luo2019minimum,atay2009mobile}. A $k$-connected network is a graph that remains connected if fewer than $k$ vertices are removed. Such systems cannot recover from a failure of $k$ or more nodes. To remedy this, a large conservative value of $k$ can be used. However, a large $k$ greatly reduces robot maneuverability, which hurts the performance of the optimization task. 

\begin{figure}[!ht]
\centering
\includegraphics[width=0.80\linewidth]{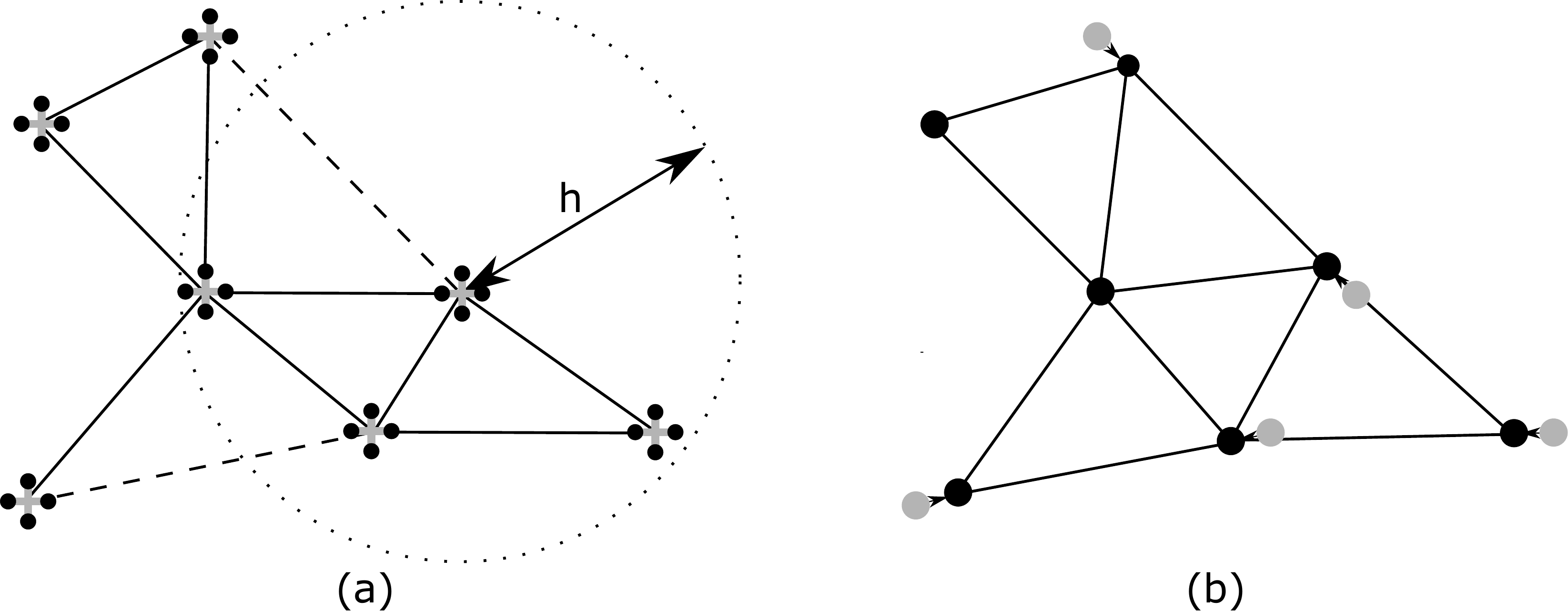}
\caption{(a) Solid lines show the communication network induced by the communication radius, $h$. Dashed lines show the edges to be augmented according to the GTO problem. (b) Hollow and solid circles refer to previous and new positions of the robots, respectively, according to the MM problem.}
\label{figintro}
\end{figure}

To address the above limitation, let us consider a generic system outlined as follows. The system operates in two modes, working mode and repair mode. In working mode, the robots perform the optimization task while always maintaining a biconnected (i.e, $k$-connected with $k=2$) topology. By using $k=2$, the system ensures high robot maneuverability and also guarantees connectedness in the case of failure of one robot. If some robots fail and the topology is no longer biconnected (but still a connected one), the system enters repair mode. Note that a biconnected network may still be connected after a failure of multiple robots. In repair mode, all the robots stop performing the optimization task and move to new positions such that biconnectivity is restored as fast as possible. When biconnectivity is achieved, the system reenters working mode. We call this \textit{Dual Mode} (\textit{DM}) framework.


In this paper, we solve the problem faced by the robots during the repair mode. The input to this problem is a set of robot positions that form a connected (but not biconnected) network. The goal is to find new positions of the robots which form a biconnected topology while minimizing the maximum distance between the previous and new positions of the robots. We call this problem the \textit{Fast Biconnectivity Restoration} (FBR) problem. Several variants of the FBR problem have been well-studied in literature~\cite{basu2004movement,abbasi2008movement,liu2010simple,lee2015connectivity}.

We develop a \textit{Quadratically Constrained Program} (QCP) formulation of the FBR problem, which is based on the idea of multi-commodity network flow~\cite{ilpref}. The QCP formulation enables us to solve the FBR problem optimally using a QCP solver. However, the QCP formulation can be used to solve only small instances of the FBR problem due to high computational overhead. We call the QCP formulation-based optimal algorithm the OPT algorithm.

We also propose an approximation algorithm to solve the FBR problem which requires less computational time than the OPT algorithm. The approximation algorithm solves the FBR problem by dividing it into two phases. In the first phase, we find a set of edges that, if augmented to the input network, make the network biconnected, such that the maximum cost of the edges is minimized. Here the cost of an edge is related to the distance between the two corresponding robots. We call this \textit{Graph Topology Optimization} (GTO) problem. In the second phase, we move the robots in such a way that the topology obtained from the first phase is realized, and the maximum distance traveled by a robot is minimized. We call this \textit{Movement Minimization} (MM) problem. One intuitive example of the GTO and MM problems is given in Figure~\ref{figintro}.

We solve the GTO problem by extending the work on the classic graph augmentation problem~\cite{eswaran1976augmentation,frederickson1981approximation} by Khuller et al.~\cite{khuller1993approximation} by considering the maximum edge cost rather than the sum of all costs. 
We call the resulting algorithm the \textit{Edge Augmentation} (EA) algorithm.
To solve the MM problem, we propose the \textit{Sequential Cascaded Relocation} (SCR) algorithm, which uses a Breadth First Search (BFS) based idea called cascaded relocation introduced in~\cite{abbasi2008movement}. We also propose a QCP formulation of the MM problem, which can be used to solve the MM problem optimally. These two solutions of the MM problem yield two algorithms to solve the FBR problem. 

We conduct extensive experimentation to evaluate the performance of our proposed algorithms. The empirical results show that our algorithms significantly outperform existing algorithms~\cite{basu2004movement,abbasi2008movement} in terms of optimizing the FBR objective, and the running time of EA-SCR is comparable to~\cite{basu2004movement,abbasi2008movement}. 

In summary, we make the following contributions:
\begin{itemize}
    \item
    {To the best of the authors' knowledge, we develop the first QCP formulation of the FBR problem which can be used to solve the FBR problem optimally.}
    \item 
    We propose approximation algorithms, EA-SCR and EA-OPT, for the FBR problem which outperform the state-of-the-art solutions. 
    \item
    We conduct experiments to compare the performance of our proposed algorithms with the existing solutions.
\end{itemize}
    

\section{Related Work}
\label{rel_work}

In a broader sense, this work is related to the connectivity maintenance problem in networked multi-robot systems. A large portion of research in this field uses algebraic graph theory to maintain the connectivity of the robotic network \cite{sabattini2013decentralized,robuffo2013passivity,zavlanos2011graph}. Sabattini et al. \cite{minelli2020self, panerati2019robust} consider both connectivity maintenance and the robustness to failures of the network by introducing extra terms into the control laws. However, they cannot guarantee the connectivity of the network after losing one robot and no repair method is provided after failures, which is the focus of our work.

One candidate approach to ensure connectivity after losing some robots is to always maintain a $k$-connected topology~\cite{luo2019minimum,atay2009mobile,cornejo2010fault}. In this approach, a larger $k$ means higher robustness to failures. However, these methods have no way of recovering from failures of $k$ or more robots, as they only focus on connectivity maintenance. Furthermore, a higher $k$ may limit the movement of the robots since it will force the robots to stay closer to each other. Recent work \cite{varadharajan2020swarm} considers the recovery plan for ground-and-air robotic networks but confines their discussion to simple topologies (i.e, a chain in their paper). By contrast, we consider the scenario in which all robots are moving, which implies a complex network structure. We try to maintain a 2-connected network and the emphasis of this paper is on how to repair the network once a robot fails. Other related works along this line \cite{hung2019hierarchical,stephan2017concurrent} emphasize more on the control system structure rather than how to recover.

There are some works from the wireless communication community that addresses failure recovery. The works most closely related to the FBR problem are conducted by Basu et al.~\cite{basu2004movement} and Abbasi et al.~\cite{abbasi2008movement}. In both works, the authors propose algorithms to make a connected network biconnected but optimize a different objective function. In~\cite{basu2004movement}, the objective is to minimize the sum of the movements of all the robots. { In~\cite{abbasi2008movement}, only a subset of the robots participate in repairing the network, and the goal is to minimize the sum of movement distances of robots involved in the repair process while our objective is to minimize the maximum movement among all the robots. Moreover, as pointed out in \cite{wang2010movement}, algorithms proposed in \cite{abbasi2008movement} may return a solution that is not biconnected in some cases. By contrast, our algorithm can guarantee biconnectivity after restoration. Similar to \cite{abbasi2008movement}, algorithms proposed in \cite{liu2010simple} can build a biconnected graph from a disconnected graph under some topological assumption, which enforces constraints between the number of robots as well as robots' initial configurations and the geometry of the environment. Lee et al.~\cite{lee2015connectivity} consider the same objective as us, but it is assumed that additional robots can be used to facilitate the transition to a biconnected topology.} 

{Our work is also closely related to \cite{luo2020minimally} and \cite{ghedini2018decentralized}. In \cite{luo2020minimally}, authors consider a more general problem on augmenting the network to be $k$-connected. In \cite{luo2020minimally}, the objective is to minimize the sum of the deviations from a prescribed controller, which doesn't necessarily imply that the maximum movement will be minimized. The algorithm in \cite{luo2020minimally} minimizes the number of edges to be augmented while our method minimizes the weight of the costliest edge to be augmented. Ghedini et al. \cite{ghedini2018decentralized} increase the resilience of network by iteratively improving upon a self-defined vulnerability metric. Their method is evaluated through experiments and it is not guaranteed that the final network is biconnected. Also, the maximum movement of robots is not considered in \cite{ghedini2018decentralized}.}

\section{Problem Formulation }
\label{pform}

First, we introduce some graph-theoretic concepts. A graph $G$ is a tuple $(V_G, E_G)$, where $V_G$ and $E_G$ are the set of vertices and edges of $G$ respectively. A graph $G$ is \textit{connected}, if there exists a path between each pair of vertices in $G$. A graph $G$ is \textit{biconnected}, if for each $v \in V_G$, the graph $G-v$ is connected. Here, $G-v$ is the graph obtained by removing from $G$ the vertex $v$ and all edges incident on $v$. A graph $G$ is \textit{barely connected}, if $G$ is connected, but not biconnected.

We assume that $n$ robots are operating in an unobstructed 2D or 3D environment. Let $\mathcal{R}=\{r_1, \ldots, r_n\}$ be the set of all robots. We denote the position of the robot $r_i$ by $x_i$, where $x_i$ is a 2D or 3D point. $X$ denotes the set of positions of the robots, $X=\{x_1,x_2,\ldots, x_n\}$.  Communication graph at positions $X$ is specified using proximity graph, i.e., a robot corresponds to a vertex $i \in {V}$ and $(i,j) \in {E}$ if $\norm{x_i-x_j} \leq h$, where $h$ is the communication radius. With slight abuse of notation, we use $G(X)$ to denote the communication graph induced by $X$. Robots have single-integrator dynamics and operate  at  maximum  speeds. Therefore, the time that it takes to transit between two points is equivalent to the corresponding distance. In the rest of this paper, we will use the maximum moving distance to denote the maximum transition time. {It should be noted that we abstract away collision avoidance from the formulation due to the fact that the communication radius of robots is usually much larger than the size of robots \cite{kuzminykh2017testing, iordache2017field} and they need to augment the network to be 2-connected only when they are already quite far from each other. In such cases, the influence of behavior controllers for collision avoidance can be ignored.}

\begin{problem}[Fast Biconnectivity Restoration~(FBR)]\label{problem: FBR}
Given current positions of robots $X_0=\{x_1, \ldots, x_n\}$, which induces a connected graph $G({X_0})$ w.r.t. communication radius $h$, find new positions of robots $X^*=\{x_1^*, \ldots, x_n^*\}$ such that $X^*$ induces a 2-connected graph and the maximum moving distance among all the robots is minimized. Mathematically,
\begin{equation}
    \begin{aligned}
        & \min_{X^*} \max_{r_i \in \mathcal{R}} \norm{x_i^* - x_i} \\
        \rm{s.t.}&~ G(X_0)~ \text{is connected},~G(X^*)~ \text{is 2-connected} .
    \end{aligned}
\end{equation}
\end{problem}

Next, we give a QCP-based formulation, which can be used to find the optimal solution to Problem \ref{problem: FBR}. 
The key idea behind the QCP formulation is based on multi-commodity network flow~\cite{ilpref}. We make use of the fact that if a graph is biconnected, there exist at least two vertex-disjoint paths between each pair of vertices. We use the following variables in the QCP formulation.

\begin{itemize}
    \item Binary variable $e_{i,j}$, where $1\leq i,j\leq n$, indicates if there exists an edge between robots $x_i^*$ and $x_j^*$ in the communication graph of $X^*$ with respect to communication radius $h$, i.e., if $x_i^*$ and $x_j^*$ are within distance $h$ of each other. 
    
    \item Binary variable $z_{s,d,i,j}$, where $1\leq s,d,i,j\leq n, \, \, s\neq d$, indicates if the edge between robots $r_i$ and $r_j$ is included in a vertex-disjoint path from source robot $r_s$ to destination robot $r_d$. 
    
    \item $x_i^*$, where $1\leq i\leq n$, represents a tuple of real-valued variables (x, y, and optionally z coordinates), which indicates the new position of the $i^{th}$ robot.
    
    \item Real valued variable $z^*$ is used to find the maximum movement of the robots.
    
\end{itemize}

The QCP formulation is presented below. We use the objective function and the set of constraints in Equation (\ref{ilpc1}) to minimize the maximum movement of the robots. Constraints (\ref{ilpc2}) ensure that if a pair of robots are within the communication radius of each other, the corresponding $e$ variable is set to 1, and vice versa. Here, M is a large positive constant. Note that, we manually set $e_{i, i}=0$, for $1\leq i\leq n$, to indicate that there are no self edges. Constraints (\ref{ilpc3}) enforce that only valid edges (i.e., edges between robots within each other's communication radius) are used to form source-destination paths. Constraints (\ref{ilpc4}) are used to maintain flow conservation. In other words, for each source-destination pair, the outgoing flow should be 2 greater than the incoming flow for the source vertex, the outgoing flow should be 2 less than the incoming flow for the destination vertex, and incoming and outgoing flow should be equal for all the other vertices. Constraints (\ref{ilpc5}) ensure vertex-disjointedness, i.e., for each source-destination pair, an internal node is included in at most one path.

$$\min  z^* $$
\begin{equation}\tag{4}
\label{ilpc1}
\text{s.t.} \norm{x_i^*-x_i}^2 \leq z^* \quad \quad  \forall \,  \, 1 \leq i \leq n 
\end{equation}  

\vspace{-0.4cm}

\begin{equation}\tag{5}
\label{ilpc2}
-{\rm M} e_{i,j} \leq \norm{x_i^*-x_j^*}^2 - h^2 \leq {\rm M}(1-e_{i,j})  \quad  \forall \, \, 1 \leq i \neq j \leq n 
\end{equation}

\vspace{-0.3cm}

\begin{equation}\tag{6}
\label{ilpc3}
z_{s,d,i,j} \leq e_{i,j}  \quad \quad \forall  \, 1 \leq s,d,i,j \leq n 
\end{equation}

\vspace{0.1cm}

$\forall \, \, 1 \leq s,d,i \leq n, \, \, \, {\rm such \, \, that,} \, s \neq d$,
\vspace{-0.1cm}
\begin{equation}\tag{7}
\label{ilpc4}
    \sum_{1 \leq j \leq n} z_{s,d,i,j} - \sum_{1 \leq j \leq n} z_{s,d,j,i} = 
\begin{cases}
    2, \; {\rm if} \, \, i = s\\
    -2, \; {\rm if} \, \, i = d\\
    0, \; {\rm otherwise} 
\end{cases}
\end{equation}

$\forall \, \, 1 \leq s,d,i \leq n, \, \, \, {\rm such \, \, that,} \, i \neq s, i \neq d, s \neq d$,
\begin{equation}\tag{8}
\vspace{-0.1cm}
\label{ilpc5}
    \sum_{1 \leq j \leq n} z_{s,d,i,j} \leq 1
\end{equation}

Note that, the QCP formulation has O($n^4$) binary variables. Hence it can be used to solve only very small instances of the FBR problem. To this end, in the following sections, we present algorithms that find approximate solutions to the FBR problem with less computational overhead.  

\section{Heuristic Algorithms}

We solve the FBR problem in a centralized way by dividing it into two sub-problems, namely, the Graph Topology Optimization problem, and the Movement Minimization problem. We define the subproblems as follows.

\begin{problem}[Graph Topology Optimization]
We are given a set of robot positions $X$, and the communication radius $h$, such that $G(X)$ is barely connected. The GTO problem aims to determine a set of edges $E_a \subseteq E_{K(X)} \backslash E_{G(X)}$, such that the graph ($V_{G(X)}, E_a \cup E_{G(X)}$) is biconnected and the weight of the most costly edge in $E_a$ is minimized, where $K(X)$ is a fully connected weighted graph induced by $X$ w.r.t. $h$. The weight of an edge between robots $r_i$ and $r_j$ in $K(X)$ is denoted by $w(i,j)$, where $w(i,j)=\max(\norm{x_i-x_j}-h,0)$.  
\end{problem}

Here, $A \backslash B$ denotes the set of elements in $A$ that are not in $B$. We call $E_a$ the \textit{Augmentation Set}, because augmenting $E_a$ to $G(X)$ makes $G(X)$ biconnected. The weight of an edge in $K(X)$ is a measure of the time required to connect the edge. Also, the edges in $G(X)$ have a weight of 0 in $K(X)$. Note that, the GTO problem aims to find the edges that need to be augmented to make $G(X)$ biconnected. The MM problem is to determine how the robots should move minimally to realize those edges.

\begin{problem}[Movement Minimization]  Given a set of positions $X$ of $n$ robots, the communication radius $h$, and the augmentation set $E_a$, where $E_a \cap E_{G(X)}=\emptyset$, the goal of the MM problem is to find new positions $X^*=\{x_1^*,x_2^*,\ldots ,x_n^*\}$ of the robots such that $E_a \cup E_{G(X)} \subseteq E_{G(X^*)}$ and the maximum movement of the robots is minimized. Mathematically,

\begin{align*} 
    & \min \max_{1 \leq i \leq n}  \norm{x_i^*-x_i} 
    \\ 
    \rm{s.t.} & \norm{x_i^*-x_j^*} \leq h, \quad \forall (i,j) \in E_a \cup E_{G(X)}.
\end{align*}
\end{problem}

In other words, the MM problem aims to move the robots in such a way that the existing edges of $G(X)$ are retained, and additionally, each pair of robots in $E_a$ comes within the communication radius of each other, while the maximum movement of the robots is minimized.

\section{Graph Topology Optimization}
\label{gtosec}

In this section, we present the \textit{Edge Augmentation} (EA) algorithm (Section~\ref{eaalgo}) after a brief overview of block-cut trees (Section~\ref{bctree})

\subsection{Block-Cut Tree}
\label{bctree}

The Block-Cut tree (BC-Tree) of a connected graph contains two types of vertices, namely, cut vertices and blocks, which we define as follows. A \textit{cut vertex} of a connected graph $G$ is a vertex that, if removed, renders $G$ disconnected. A \textit{block} of a connected graph $G$ is a maximal connected subgraph of $G$ that has no cut vertex. It should be noted that a biconnected graph itself is one block.

\begin{figure}[!ht]
\centering
\includegraphics[width=1.0\linewidth]{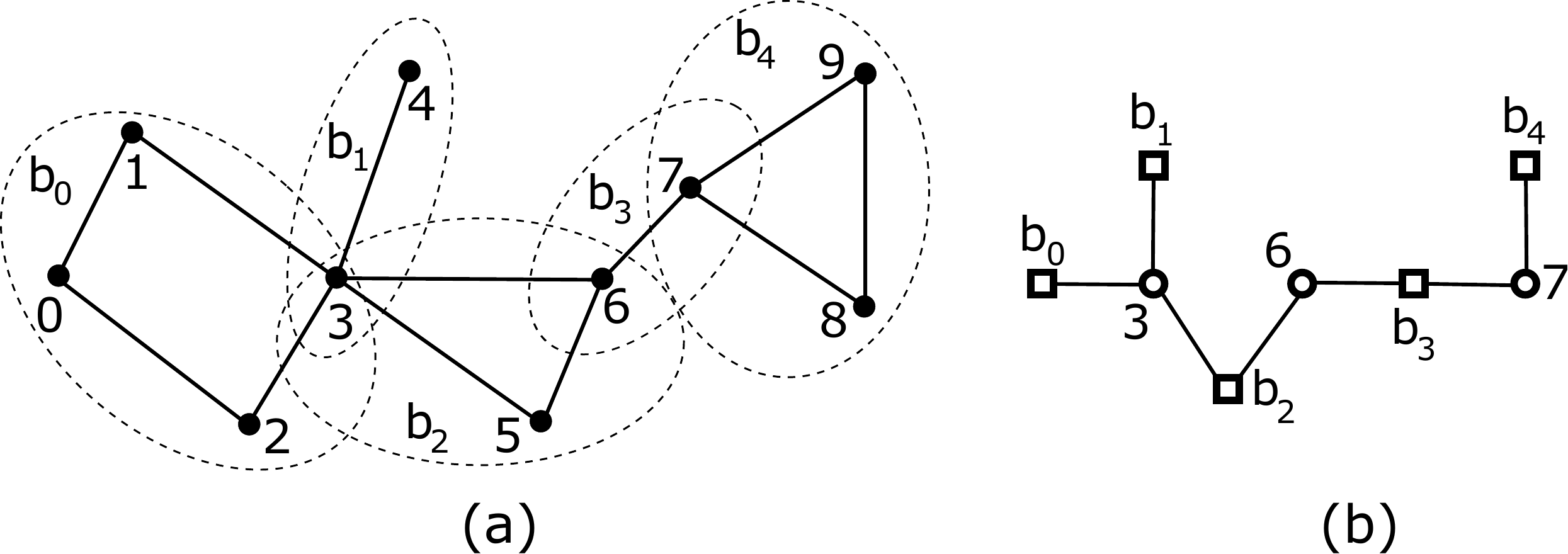}
\caption{(a) Dotted ovals show the blocks. (b) The corresponding BC-Tree. Hollow squares and circles show blocks and cut vertices respectively.}
\label{bctfig}
\end{figure}

Now we describe the construction of the BC-Tree of a connected graph $G$ according to \cite{west2001introduction}.  We denote the set of blocks and cut vertices of $G$ by $B$ and $C$ respectively. Let $V(b)$ denote the set of vertices of block $b$. A BC-Tree of $G$ is a tree whose vertex $v$ is either a block $b \in B$ or a cut vertex $c\in C$. Edges only exist between different types of vertices. An edge exists between $b \in B$ and $c \in C$ if and only if $c \in V(b)$, i.e., $b$ includes $c$. An illustrative example of a BC-Tree is given in Figure~\ref{bctfig}.

\subsection{Edge Augmentation Algorithm}
\label{eaalgo}

We propose the EA algorithm (Algorithm \ref{algorithm:biconnectivity_augmentation}) to solve the GTO problem. The algorithm is inspired by \cite{khuller1993approximation} but the key difference is that \cite{khuller1993approximation} concerns the sum of the costs of the edges, while we care about the maximum cost of the edges. 
The EA algorithm takes as input a set of robot positions $X$ and the communication radius $h$. We assume that the $G(X)$ is barely connected. The objective of the EA algorithm is to select a set of edges, $E_a$, which are not present in $G(X)$ and augmentation of which makes $G(X)$ biconnected, such that the maximum weight of the edges of $E_a$ is minimized. 

\begin{algorithm}[ht]\label{algorithm:biconnectivity_augmentation}
    \caption{Edge Augmentation}
    \SetKwInOut{Input}{Input}
    \SetKwInOut{Output}{Output}
    \underline{function EA} $(X, h)$ \\
    \Input{
    \begin{itemize}
        \item The set of robot positions, $X$ 
        \item The communication radius, $h$
    \end{itemize}
    }
    \Output{Augmentation set, $E_a$}
    $T \gets {\rm ConstructBCTree}(G(X))$ \\
    $E_{a}, \, E_{c} \gets \emptyset, \, E_{K(X)} \backslash E_{G(X)}$ \\
    Superimpose all edges in $E_c$ on $T$. Discard the self-loops. Keep only the cheapest edge between two nodes in $T$. \\
    
    Initialize a directed null graph $G_D=(V_T, \emptyset)$. \\
    
    Choose an arbitrary leaf of $T$ as root $r_l$, and direct all the edges of $T$ towards $r_l$. The resulting directed tree is still denoted by $T=(V_T, E_T)$. \\
    
    Add all edges in $E_T$ to $G_D$ and set their cost to 0. \\
    
    For each superimposed edge, generate the \textit{image edges} according to Routine 2(c) of Algorithm \textit{Find Vertex Aug.} in~\cite{khuller1993approximation} and add them to $G_D$. \\
    
    Find an MBSA $G_M$ of $G_D$ rooted at $r_l$.
    Identify edges $E$ of $G_M$ that are not in $T$, and add to $E_a$ the corresponding edges in $E_c$ that generated $E$. \\
    Return $E_a$.
\end{algorithm}


In the EA algorithm, first, we construct the BC-Tree, $T$, of $G(X)$ (Line 2) and initialize the set of candidate edges, $E_c$, as the set of all the edges that are not in $G(X)$ (Line 3). Next, we \textit{superimpose} all the candidate edges on $T$ (Line 4). Superimposing an edge $(i,j) \in E_c$ on $T$ involves finding the corresponding vertices of robot $r_i$ and $r_j$ on $T$ and adding an edge with cost $w(i,j)$ between the corresponding vertices.  Next, we make $T$ directed towards an arbitrary leaf node $r_l$ (Line 6), and add all the directed edges of $T$ to a directed graph $G_D$ (Line 7). We generate \textit{image}s of the edges superimposed on $T$ using Routine 2(c) of Algorithm \textit{Find Vertex Aug.} in~\cite{khuller1993approximation} and add the image edges to $G_D$ (Line 8). Next, we construct the Minimum Bottleneck Spanning Arborescence (MBSA) $G_M$ of $G_D$ rooted at $r_l$. The MBSA is a directed spanning tree where the most expensive edge is as cheap as possible. We identify the edges in $G_M$ that are not in $T$ and add the corresponding candidate edges to $E_a$ (Line 9). The EA algorithm differs from \cite{khuller1993approximation} only in Line 9, where we compute an MBSA instead of a directed MST. 

\begin{theorem}\label{theorem: biconnect_guarantee}
The augmentation set returned by Algorithm \ref{algorithm:biconnectivity_augmentation} biconnects $G(X)$.
\end{theorem}
The proof of Theorem \ref{theorem: biconnect_guarantee}is similar to Lemma 4.6 in \cite{khuller1993approximation} and is omitted here for brevity. Note that the number of edges in the augmentation set is not necessarily minimum since the objective in this work is related to only the edge cost.



\section{Movement Minimization}
\label{mmsec}

The MM problem aims to move the robots to new positions such that the edges returned by the EA algorithm are realized. The input to the MM problem is the set of robot positions $X$, $h$, and the augmentation set $E_a$ which contains edges that are not in $G(X)$. The MM problem aims to move the robots such that the existing edges of $G(X)$ are retained, and each pair of robots in $E_a$ becomes directly connected, while the maximum movement of the robots is minimized. We propose two algorithms for the MM problem as follows.

\subsection{Sequential Cascaded Relocation Algorithm}
\label{scrsec}

We propose the \textit{Sequential Cascaded Relocation} (SCR) algorithm which is based on the idea of cascaded relocation introduced by Abbasi et al.~\cite{abbasi2008movement}. First, we describe the process of cascaded relocation as proposed in~\cite{abbasi2008movement}. A cascaded relocation refers to moving one vertex of a graph to a new location while retaining the existing edges of the graph. Note that, moving only one vertex may disconnect the existing edges of $G(X)$. To remedy this, a Breadth-First Search (BFS) like approach is employed in~\cite{abbasi2008movement} so that existing edges in $G(X)$ are retained. Let $r$ be the vertex that is to be relocated. We consider the neighbors of $r$ in increasing order of graph distance from $r$. First, we relocate $r$ to the desired position. At iteration $d$, we consider all the $d$-hop neighbors of $r$ in $G(X)$. If a neighbor disconnects from its parent due to the relocation of the parent, we move the neighbor minimally towards the parent such that the disconnected edge is reestablished. Thus each relocation may result in a cascade of relocations so that the existing edges of the communication graph are retained.

Now we describe the SCR algorithm, where we augment the edges in $E_a$ to the communication graph by performing a sequence of cascaded relocations. In the SCR algorithm, for each edge $(i,j) \in E_a$, we make two cascaded relocations. First, we move the robot $r_i$ towards $r_j$ by a distance of $\frac{1}{2}w(i,j)$. Next, we move robot $r_j$ towards $r_i$ minimally such that $r_i$ and $r_j$ are within distance $h$ of each other. Each pair of relocations establishes one edge of $E_a$.

\subsection{QCP-based Formulation for MM}
\label{optsec}

We propose a QCP formulation of the MM problem to solve the problem optimally. The QCP of the MM problem is presented below. Here, the variables $x_i^*$ and $z^*$ serve the same purpose as the QCP formulation in Section~\ref{pform}.
\begin{align*}
    & \min ~~ z^* \\
    \rm{s.t.}~&~ \norm{x_i^*-x_i}^2 \leq z^* \quad \quad \forall  \, 1 \leq i \leq n \\
    &~\norm{x_i^*-x_j^*}^2 \leq h^2 \quad \quad \forall \, (i,j) \in E_a \cup E_{G(X)}
\end{align*}

\section{Experiments}
\label{exp}


\subsection{Experimental Setup}
\label{expsetup}

\textbf{Evaluation Metric}: We use two metrics to empirically evaluate our proposed algorithms: minmax distance and running time. The objective of the FBR problem is to minimize the maximum distance between the previous and new positions of the robots. We call this objective the minmax distance. 

\textbf{Compared Algorithms}: We empirically compare the performance of three algorithms proposed in this paper (OPT, EA-SCR, EA-OPT) and two existing algorithms (BT~\cite{basu2004movement}, CR~\cite{abbasi2008movement}). OPT is described in Section~\ref{pform}. EA-SCR is described in Section~\ref{eaalgo} and~\ref{scrsec}. EA-OPT is described in Section~\ref{eaalgo} and~\ref{optsec}. We use a commercial QCP solver, Gurobi~\cite{grb}, to solve the QCPs in OPT and EA-OPT.

The algorithm presented in~\cite{basu2004movement} is based on BC-Tree. In each iteration of this algorithm, the block of the BC-Tree with the maximum number of vertices is selected as the root node. Next, each leaf block of the BC-Tree rooted at the root node is translated towards its parent node such that the leaf block merges with the parent block. The BC-Tree is reconstructed after the translation. This process is repeated until there is one block in the BC-Tree, i.e., the graph is biconnected. We call this \textit{Block Translation} (BT) algorithm.

In the case of the algorithm presented in~\cite{abbasi2008movement}, the input is a connected graph that results from the removal of a vertex from a biconnected graph, and the position of the removed vertex. The goal of this algorithm is to move a minimum number of nodes to make the graph biconnected. In this algorithm, first, the neighbor of the removed vertex with the lowest degree is selected as the \textit{Best Neighbor} (BN) node. The BN node is moved towards the position of the removed vertex until connections of the BN node with all the neighbors of the removed vertex are established. The movement of the BN node may disconnect some of its previous edges, which is fixed by performing one cascaded relocation (as described in Section~\ref{scrsec}) for the BN node. We modify this algorithm as follows. We select the closest neighbor of the removed vertex as the BN node, instead of selecting the neighbor with the lowest degree. We call this modified algorithm \textit{Cascaded Relocation} (CR) algorithm.

\textbf{Dataset}: We construct a synthetic dataset as follows. First, we generate random 2D points representing the position of the robots. Next, we construct the communication graph of the generated points using $h=1m$. If the communication graph is biconnected, we remove one point and reconstruct the communication graph. If the new graph is not biconnected, we add the set of points to our dataset. Thus, we ensure our dataset contains only barely connected graphs.  

\textbf{Platform}: The algorithms are implemented using C++. The experiments are conducted on a core-i7 2GHz PC with 8GB RAM, running Microsoft Windows 10.

\subsection{A Qualitative Example}
\label{qualex}

\begin{figure}[!ht]
\centering
\includegraphics[width=0.58\linewidth]{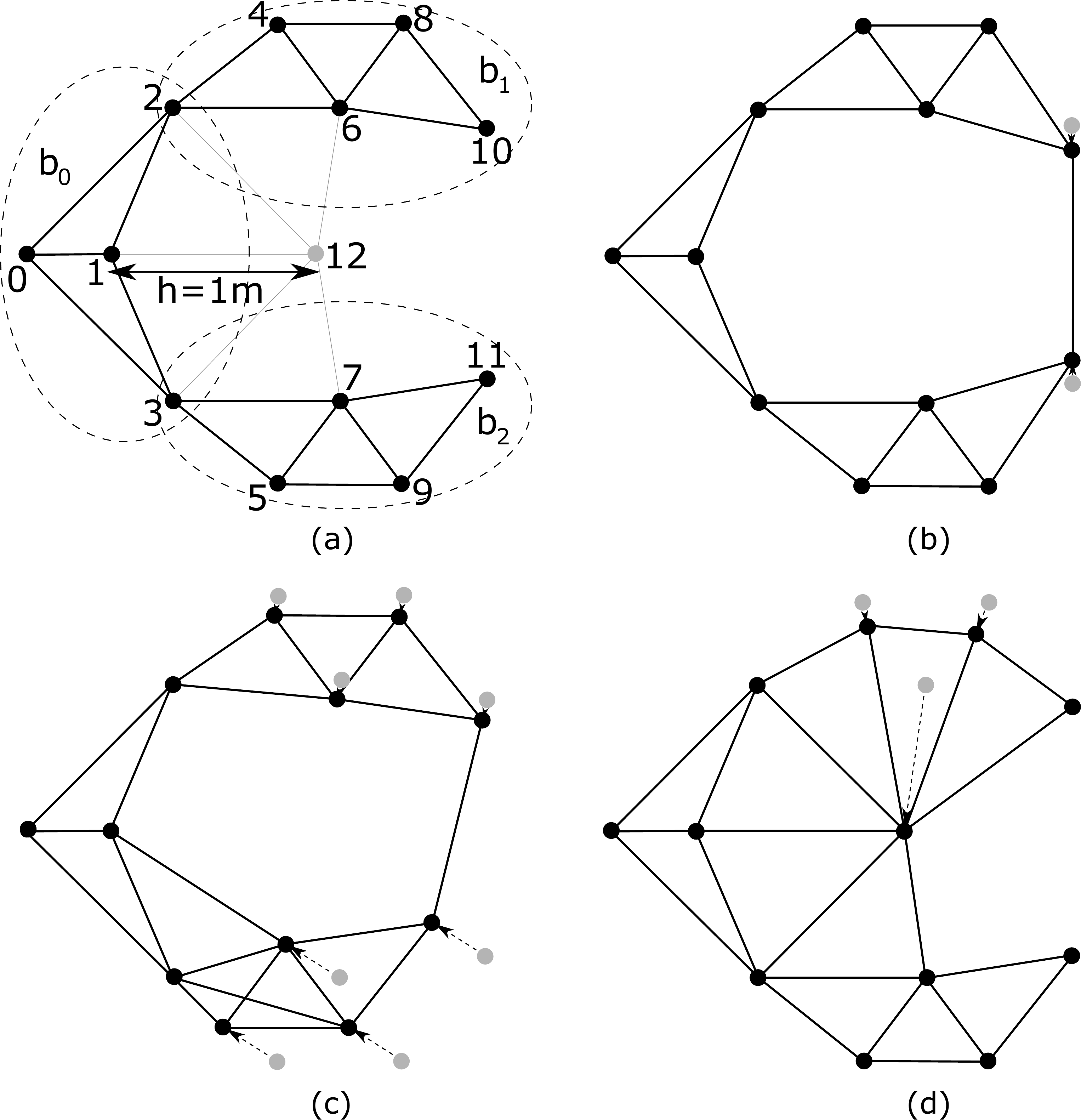}
\caption{(a) A barely connected graph created by removing vertex 12 from a biconnected graph. Dotted ovals show the blocks. (b), (c), and (d) show the outputs of the EA-SCR, BT, and CR algorithms respectively. The corresponding minmax distances are 0.1m, 0.3m, and 0.71m respectively.}
\label{figexp}
\end{figure}

In this section, we explain with an example how our proposed algorithms achieve a lower minmax distance than the existing solutions using the example shown in Figure~\ref{figexp}(a). 

In the case of our proposed EA-SCR and EA-OPT algorithms as shown in Figure~\ref{figexp}(b), during the first phase, the edge between vertices 10 and 11 is added to the augmentation set. In the second phase, vertices 10 and 11 move minimally towards each other until an edge between them is established.

In the case of BT algorithm as shown in Figure~\ref{figexp}(c), in the first iteration, block $b_1$ is selected as the root block, and leaf block $b_2$ is moved to the left to merge with its parent block $b_0$. In the next iteration, the merged block becomes the root, and hence $b_1$ is moved downwards to form a biconnected graph. Thus, in BT algorithm, block $b_2$ moves unnecessarily to the left, which increases the minmax distance.

In the case of CR algorithm, as shown in Figure~\ref{figexp}(d), vertex 6 is selected as the BN node, and it is moved towards the deleted node 12. Note that, vertex 6 needs to move significantly downwards to establish an edge with vertex 1, which results in a high minmax distance.

\subsection{Empirical Results}
\label{empres}

\begin{figure}[!ht]
\centering
\includegraphics[width=0.70\linewidth]{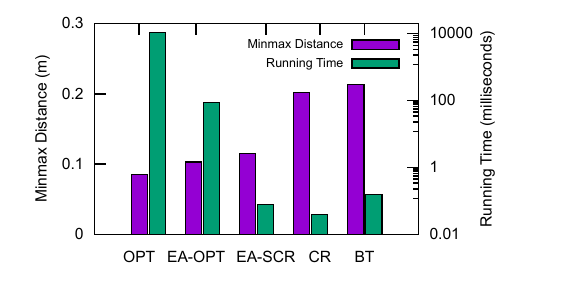}
\caption{Comparison with OPT algorithm using $n=8$ robots.}
\label{figc}
\end{figure}

\textbf{Comparison with OPT}: In this experiment, we compare the minmax distance and running time of the discussed algorithms. As the OPT algorithm can be used to solve only small instances of the FBR problem, we use 8 robots in this experiment. The experimental results in Figure~\ref{figc} show that the OPT algorithm requires significantly higher running time than the other algorithms (because of the use of the QCP solver). In terms of minmax distance, our proposed algorithms (EA-SCR and EA-OPT) perform close to the OPT algorithm and significantly better than the existing algorithms (CR and BT). In all the experiments, we use a communication radius of $1m$. We perform each experiment 100 times and report the average.

\begin{figure}[!ht]
\centering
\includegraphics[width=0.70\linewidth]{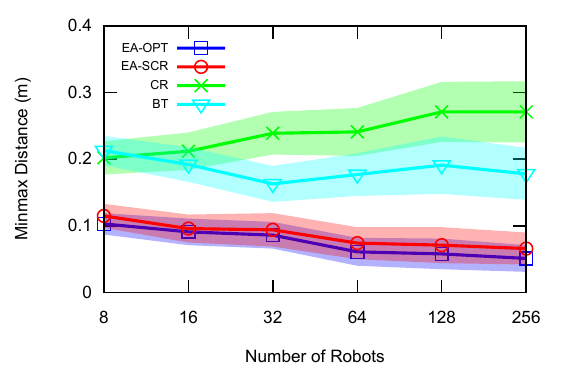}
\caption{Comparison of minmax distance.}
\label{figd}
\end{figure}

\textbf{Comparison of Minmax Distance}: In this experiment, we compare the discussed algorithms (except for OPT) in terms of minmax distance. We vary the number of robots from 8 to 256 by a factor of 2. The experimental results in Figure~\ref{figd} show that our proposed algorithms (EA-OPT and EA-SCR) perform better than the existing algorithms (CR and BT). The standard deviation is shown using the shades.

\begin{figure}[!ht]
\centering
\includegraphics[width=0.70\linewidth]{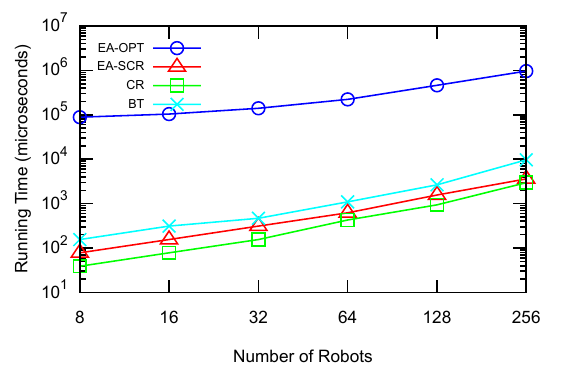}
\caption{Comparison of running time.}
\label{figt}
\end{figure}

\textbf{Comparison of Running Time}: To compare the running time of the proposed algorithms, we use the same setup as the previous experiment. The experimental results in Figure~\ref{figt} show that the EA-OPT algorithm requires the highest running time (because of the use of the QCP solver) and the other three algorithms require similar computational time.

EA-SCR presents the best trade-off between minmax distance and running time. It achieves minmax distance that is comparable to OPT and EA-OPT, and much better than CR and BT. The running time of EA-SCR is comparable to CR and BT, and significantly better than OPT and EA-OPT. 

\section{Case Study: Persistent Monitoring}
\label{cspm}

In this section, we demonstrate the applicability of our proposed algorithms in the context of a practical multi-robot optimization problem, i.e., the \textit{Persistent Monitoring} (\textit{PM}) problem. In a typical setup of the PM problem as in~\cite{rabban2021failure,shi2021communication,rabban2019mvfs}, multiple robots monitor an occluded grid-based environment. Each grid-cell in the environment has a latency value in the range $[0, l_{max}]$, which depends on the last time the cell was visible from some robot. The latency value of a cell is 0 if it is visible from some robot in the current time step. Otherwise, the latency value increases linearly at each time step, until it reaches $l_{max}$. The PM problem is to select one trajectory for each robot from a set of candidate trajectories, such that, the sum of the latency values is minimized. 

In this experiment, we simulate a $32m \times 32m$ environment discretized into $160 \times 160$ cells each of size $0.2m \times 0.2m$. There are 60 obstacles, each with a height of $2.5m$, which occupy approximately 15\% of the environment. A team of UAVs equipped with downward-facing cameras flying at a fixed height of $5m$ is performing the PM task. The communication radius, $h$, is $10m$. The visibility footprint of a UAV camera on the ground is a circle with a radius of $3m$. Each UAV has 4 candidate trajectories, $\{$forward, backward, left, and right$\}$. The UAVs move at a speed of $0.2m$ per time-step. $l_{max}$ is set to 100 and the latency of non-visible cells are set to increase by 1 unit per time-step.  

\begin{figure}[!ht]
\centering
\includegraphics[width=1.00\linewidth]{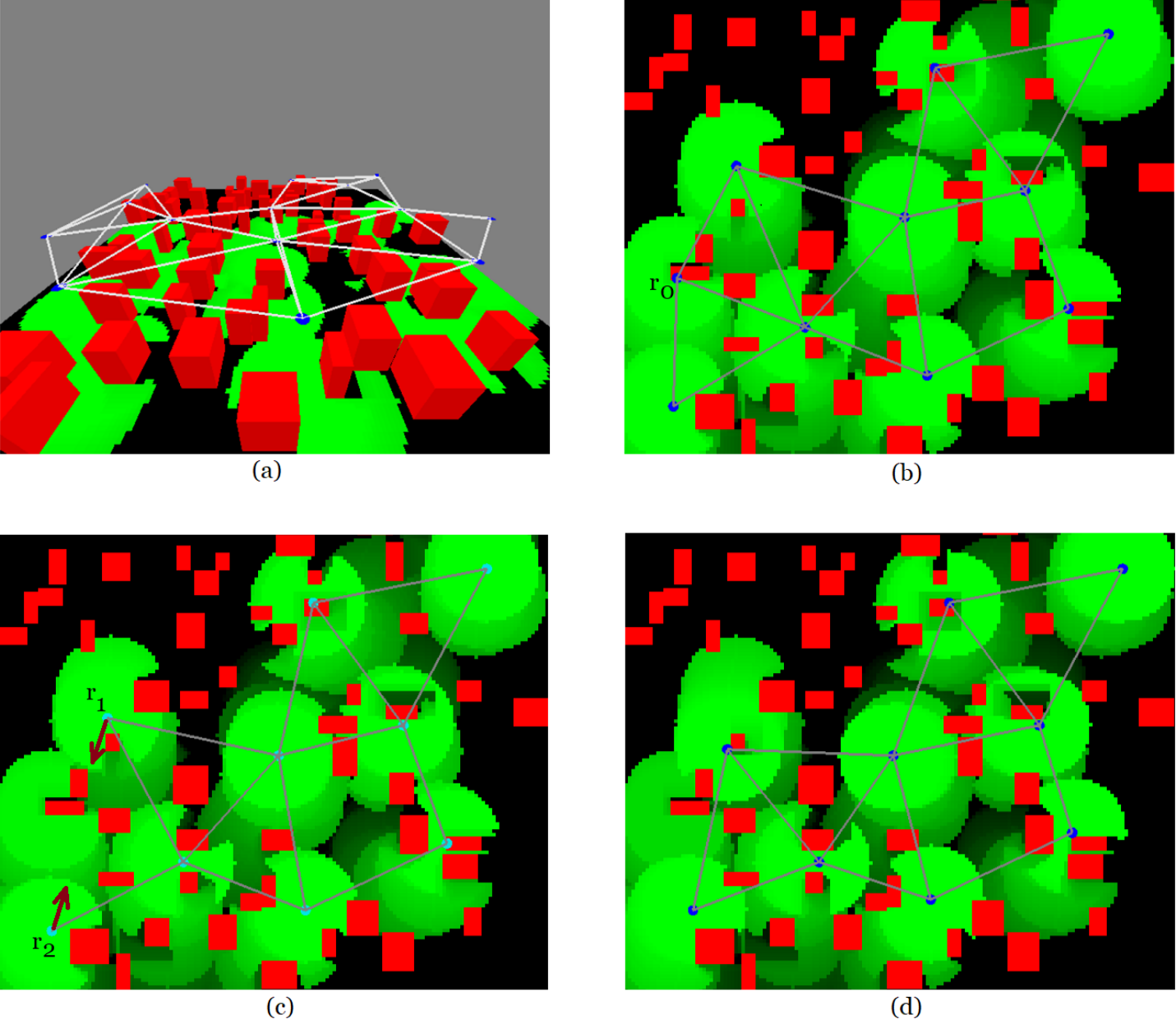}
\caption{Blue circles and red boxes represent UAVs and obstacles respectively. Shades between green and black represent latency of the cells, where green and black stands for $0$ and $l_{max}$ respectively. (a) 3D perspective view. (b) Top down view. UAV $r_0$ is about to fail. (c) Failure of $r_0$ makes the topology barely connected. UAVs $r_1$ and $r_2$ move towards each other. (d) Biconnectivity is restored.}
\label{snap}
\end{figure}

\vspace{-0.2cm}

The UAVs perform the PM task according to the DM framework introduced in Section~\ref{intro}. The UAVs continuously broadcast their locations throughout the network. Therefore, as long as the communication graph is connected, each UAV  has complete knowledge of the PM task state, i.e., locations of all other UAVs, and latency values of all the cells.


At each time-step during the working mode, each UAV individually computes a joint motion plan of the team for one time-step that greedily optimizes the PM objective. Note that, as the UAVs have complete knowledge of the PM task state, all the UAVs will come up with the same plan. If the joint motion plan does not break biconnectivity, the UAVs execute the planned motion. Otherwise, all the UAVs move in a common direction, which guarantees that biconnectivity will not be lost. Although there exists other ways to maintain biconnectivity during working mode, we use this simplistic approach for brevity and clarity of presentation.

The system enters repair mode if one UAV fails and the failure makes the network barely connected. Then each UAV individually runs the FBR routine to determine where to move to restore biconnectivity. During repair mode, a UAV moves $0.2m$ per time-step until it reaches the destination. An alternative to each UAV computing the joint motion plan and the FBR routine by itself is that a leader robot does the computation and communicates the findings to the other UAVs. Some snapshots of the system in operation are shown in Figure~\ref{snap}. The system is visualized using OpenGL\footnote{\url{http://raaslab.org/vids/rabban2021fbr.mp4}}.

\begin{figure}[!ht]
\centering
\includegraphics[width=0.85\linewidth]{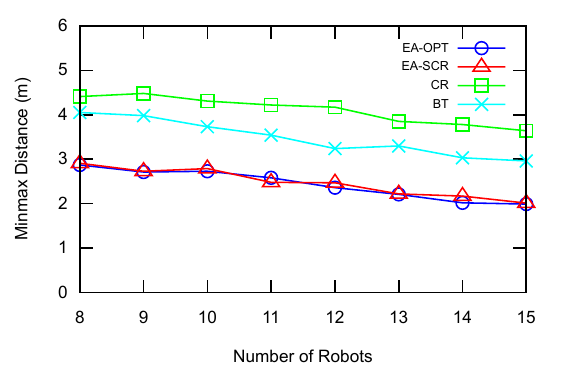}
\caption{Comparison of minmax distance in persistent monitoring task.}
\label{rpm}
\end{figure}

In the first experiment, we assume that at each time-step during the working mode, a UAV fails independently with a probability of 0.001, but the UAVs do not fail during the repair mode. The team initially consists of 16 UAVs. We run the simulation until the number of UAVs drops to 8. We call each such simulation an \textit{episode}. The experimental results presented in Figure~\ref{rpm} show the average minmax distance over 100 episodes. We observe that the performance of the algorithms is consistent with the findings of Section~\ref{exp}.

In the next experiment, we assume that each UAV fails independently with a probability of 0.001 at each time-step regardless of whether the system is in working mode or repair mode. We call the event when a UAV fails during the repair mode and thereafter makes the network disconnected a \textit{disconnection}. We call an episode a \textit{success} if the UAV team size reaches 8 without disconnection starting from an initial size of 16. If an episode does not succeed, we call the time-step number when the disconnection occurs the \textit{disconnection time}. We simulate 100 episodes and report the number of successes and the average disconnection time of unsuccessful episodes in Table~\ref{tbl}. The results show that our proposed algorithms outperform the existing ones both in maximizing success count and average disconnection time.

\begin{table}
\begin{center}
\caption{\label{tbl}}
 \begin{tabular}{|c|c|c|c|c|} 
 \hline
 Algorithm & EA-OPT & EA-SCR & CR & BT \\ 
 \hline \hline
 Success Count (out of 100) & 48 & 47 & 35 & 42 \\ 
 \hline
 Disconnection Time (timesteps) & 387 & 375 & 273 & 309 \\
 \hline
\end{tabular}
\end{center}
\end{table}
\section{Conclusion}
\label{con}

In this work, we have proposed algorithms for the FBR problem which has outperformed the existing algorithms. We have also developed a QCP formulation to solve the FBR problem optimally. We have demonstrated the effectiveness of our proposed solution by conducting empirical studies. 

In the future, we intend to prove the approximability of the performance of EA-SCR algorithm. We also plan to work on the FBR variant where the environment contains obstacles. One possible way to deal with obstacles is to discretize the environment and solve a discrete version of FBR. Another direction is to devise the distribute counterpart of the proposed algorithm. 

\bibliographystyle{IEEEtran}

@Article{rabban2021failure,
  author  = {Md. {Ishat-E-Rabban} and Pratap Tokekar},
  journal = {IEEE Robotics and Automation Letters},
  title   = {Failure-Resilient Coverage Maximization with Multiple Robots},
  year    = {2021},
  note    = {},
  url     = {https://arxiv.org/abs/2007.02204},
}

@inproceedings{wolff2012robust,
  title={Robust control of uncertain Markov decision processes with temporal logic specifications},
  author={Wolff, Eric M and Topcu, Ufuk and Murray, Richard M},
  booktitle={2012 IEEE 51st IEEE Conference on Decision and Control (CDC)},
  pages={3372--3379},
  year={2012},
  organization={IEEE}
}

@article{saulnier2017resilient,
  title={Resilient flocking for mobile robot teams},
  author={Saulnier, Kelsey and Saldana, David and Prorok, Amanda and Pappas, George J and Kumar, Vijay},
  journal={IEEE Robotics and Automation letters},
  volume={2},
  number={2},
  pages={1039--1046},
  year={2017},
  publisher={IEEE}
}

@article{guerrero2018design,
  title={Design guarantees for resilient robot formations on lattices},
  author={Guerrero-Bonilla, Luis and Saldana, David and Kumar, Vijay},
  journal={IEEE Robotics and Automation Letters},
  volume={4},
  number={1},
  pages={89--96},
  year={2018},
  publisher={IEEE}
}

@incollection{saldana2018triangular,
  title={Triangular networks for resilient formations},
  author={Saldana, David and Prorok, Amanda and Campos, Mario FM and Kumar, Vijay},
  booktitle={Distributed Autonomous Robotic Systems},
  pages={147--159},
  year={2018},
  publisher={Springer}
}

@incollection{saldana2019resilient,
  title={Resilient backbones in hexagonal robot formations},
  author={Salda{\~n}a, David and Guerrero-Bonilla, Luis and Kumar, Vijay},
  booktitle={Distributed Autonomous Robotic Systems},
  pages={427--440},
  year={2019},
  publisher={Springer}
}

@article{minelli2020self,
  title={Self-optimization of resilient topologies for fallible multi-robots},
  author={Minelli, Marco and Panerati, Jacopo and Kaufmann, Marcel and Ghedini, Cinara and Beltrame, Giovanni and Sabattini, Lorenzo},
  journal={Robotics and Auton. Systems},
  volume={124},
  pages={103384},
  year={2020},
  publisher={Elsevier}
}

@article{panerati2019robust,
  title={Robust connectivity maintenance for fallible robots},
  author={Panerati, Jacopo and Minelli, Marco and Ghedini, Cinara and Meyer, Lucas and Kaufmann, Marcel and Sabattini, Lorenzo and Beltrame, Giovanni},
  journal={Autonomous Robots},
  volume={43},
  number={3},
  pages={769--787},
  year={2019},
  publisher={Springer}
}

@article{yang2010decentralized,
  title={Decentralized estimation and control of graph connectivity for mobile sensor networks},
  author={Yang, Peng and Freeman, Randy A and Gordon, Geoffrey J and Lynch, Kevin M and Srinivasa, Siddhartha S and Sukthankar, Rahul},
  journal={Automatica},
  volume={46},
  number={2},
  pages={390--396},
  year={2010},
  publisher={Elsevier}
}

@inproceedings{luo2019minimum,
  title={Minimum k-Connectivity Maintenance for Robust Multi-Robot Systems},
  author={Luo, Wenhao and Sycara, Katia},
  booktitle={2019 IEEE/RSJ International Conference on Intelligent Robots and Systems (IROS)},
  pages={7370--7377},
  year={2019},
  organization={IEEE}
}

@incollection{atay2009mobile,
  title={Mobile wireless sensor network connectivity repair with k-redundancy},
  author={Atay, Nuzhet and Bayazit, Burchan},
  booktitle={Algorithmic Foundation of Robotics VIII},
  pages={35--49},
  year={2009},
  publisher={Springer}
}

@INPROCEEDINGS{rama2019resilience,  author={R. K. {Ramachandran} and J. A. {Preiss} and G. S. {Sukhatme}},  booktitle={2019 IEEE/RSJ International Conference on Intelligent Robots and Systems (IROS)},   title={Resilience by Reconfiguration: Exploiting Heterogeneity in Robot Teams},   year={2019},  volume={},  number={},  pages={6518-6525},}

@article{anari2016euclidean,
  title={Euclidean movement minimization},
  author={Anari, Nima and Fazli, MohammadAmin and Ghodsi, Mohammad and Safari, MohammadAli},
  journal={Journal of Combinatorial Optimization},
  volume={32},
  number={2},
  pages={354--367},
  year={2016},
  publisher={Springer}
}

@inproceedings{chakraborty2010reconfiguration,
  title={Reconfiguration algorithms for mobile robotic networks},
  author={Chakraborty, Nilanjan and Sycara, Katia},
  booktitle={2010 IEEE International Conference on Robotics and Automation},
  pages={5484--5489},
  year={2010},
  organization={IEEE}
}

@inproceedings{engin2018minimizing,
  title={Minimizing movement to establish the connectivity of randomly deployed robots},
  author={Engin, K Selim and Isler, Volkan},
  booktitle={Twenty-Eighth International Conference on Automated Planning and Scheduling},
  year={2018}
}

@book{west2001introduction,
  title={Introduction to graph theory},
  author={West, Douglas Brent and others},
  volume={2},
  year={2001},
  publisher={Prentice hall Upper Saddle River}
}

@inproceedings{zhang2012robustness,
  title={Robustness of information diffusion algorithms to locally bounded adversaries},
  author={Zhang, Haotian and Sundaram, Shreyas},
  booktitle={2012 American Control Conference (ACC)},
  pages={5855--5861},
  year={2012},
  organization={IEEE}
}

@article{krause2008near,
  title={Near-optimal sensor placements in Gaussian processes: Theory, efficient algorithms and empirical studies},
  author={Krause, Andreas and Singh, Ajit and Guestrin, Carlos},
  journal={Journal of Machine Learning Research},
  volume={9},
  number={Feb},
  pages={235--284},
  year={2008}
}

@inproceedings{ames2019control,
  title={Control barrier functions: Theory and applications},
  author={Ames, Aaron D and Coogan, Samuel and Egerstedt, Magnus and Notomista, Gennaro and Sreenath, Koushil and Tabuada, Paulo},
  booktitle={2019 18th European Control Conference (ECC)},
  pages={3420--3431},
  year={2019},
  organization={IEEE}
}

@article{borrmann2015control,
  title={Control barrier certificates for safe swarm behavior},
  author={Borrmann, Urs and Wang, Li and Ames, Aaron D and Egerstedt, Magnus},
  journal={IFAC-PapersOnLine},
  volume={48},
  number={27},
  pages={68--73},
  year={2015},
  publisher={Elsevier}
}

@article{abbasi2008movement,
  title={Movement-assisted connectivity restoration in wireless sensor and actor networks},
  author={Abbasi, Ameer A and Younis, Mohamed and Akkaya, Kemal},
  journal={IEEE Transactions on parallel and distributed systems},
  volume={20},
  number={9},
  pages={1366--1379},
  year={2008},
  publisher={IEEE}
}

@article{basu2004movement,
  title={Movement control algorithms for realization of fault-tolerant ad hoc robot networks},
  author={Basu, Prithwish and Redi, Jason},
  journal={IEEE network},
  volume={18},
  number={4},
  pages={36--44},
  year={2004},
  publisher={IEEE}
}

@article{abbasi2012least,
  title={A least--movement topology repair algorithm for partitioned wireless sensor--actor networks},
  author={Abbasi, Ameer Ahmed and Younis, Mohamed F and Baroudi, Uthman A},
  journal={International Journal of Sensor Networks},
  volume={11},
  number={4},
  pages={250--262},
  year={2012},
  publisher={Inderscience Publishers}
}

@article{wang2010movement,
  title={On “movement-assisted connectivity restoration in wireless sensor and actor networks”},
  author={Wang, ShiGuang and Mao, XuFei and Tang, Shao-Jie and Li, XiangYang and Zhao, JiZhong and Dai, Guojun},
  journal={IEEE Transactions on Parallel and Distributed Systems},
  volume={22},
  number={4},
  pages={687--694},
  year={2010},
  publisher={IEEE}
}

@article{liu2010simple,
  title={Simple movement control algorithm for bi-connectivity in robotic sensor networks},
  author={Liu, Hai and Chu, Xiaowen and Leung, Yiu-Wing and Du, Rui},
  journal={IEEE Journal on Selected Areas in Communications},
  volume={28},
  number={7},
  pages={994--1005},
  year={2010},
  publisher={IEEE}
}

@article{lee2015connectivity,
  title={Connectivity restoration in a partitioned wireless sensor network with assured fault tolerance},
  author={Lee, Sookyoung and Younis, Mohamed and Lee, Meejeong},
  journal={Ad Hoc Networks},
  volume={24},
  pages={1--19},
  year={2015},
  publisher={Elsevier}
}

@inproceedings{butterfield2008autonomous,
  title={Autonomous biconnected networks of mobile robots},
  author={Butterfield, Jesse and Dantu, Karthik and Gerkey, Brian and Jenkins, Odest Chadwicke and Sukhatme, Gaurav S},
  booktitle={2008 6th International Symposium on Modeling and Optimization in Mobile, Ad Hoc, and Wireless Networks and Workshops},
  pages={640--646},
  year={2008},
  organization={IEEE}
}

@article{frederickson1981approximation,
  title={Approximation algorithms for several graph augmentation problems},
  author={Frederickson, Greg N and Ja’Ja’, Joseph},
  journal={SIAM Journal on Computing},
  volume={10},
  number={2},
  pages={270--283},
  year={1981},
  publisher={SIAM}
}

@article{khuller1993approximation,
  title={Approximation algorithms for graph augmentation},
  author={Khuller, Samir and Thurimella, Ramakrishna},
  journal={Journal of algorithms},
  volume={14},
  number={2},
  pages={214--225},
  year={1993},
  publisher={Elsevier}
}

@article{bienstock1993note,
  title={A note on the prize collecting traveling salesman problem},
  author={Bienstock, Daniel and Goemans, Michel X and Simchi-Levi, David and Williamson, David},
  journal={Mathematical programming},
  volume={59},
  number={1-3},
  pages={413--420},
  year={1993},
  publisher={Springer}
}

@article{goemans2009combining,
  title={Combining approximation algorithms for the prize-collecting TSP},
  author={Goemans, Michel X},
  journal={arXiv preprint arXiv:0910.0553},
  year={2009}
}

@article{balas1989prize,
  title={The prize collecting traveling salesman problem},
  author={Balas, Egon},
  journal={Networks},
  volume={19},
  number={6},
  pages={621--636},
  year={1989},
  publisher={Wiley Online Library}
}

@article{balas1995prize,
  title={The prize collecting traveling salesman problem: II. Polyhedral results},
  author={Balas, Egon},
  journal={Networks},
  volume={25},
  number={4},
  pages={199--216},
  year={1995},
  publisher={Wiley Online Library}
}

@article{kuo2014maximizing,
  title={Maximizing submodular set function with connectivity constraint: Theory and application to networks},
  author={Kuo, Tung-Wei and Lin, Kate Ching-Ju and Tsai, Ming-Jer},
  journal={IEEE/ACM Transactions on Networking},
  volume={23},
  number={2},
  pages={533--546},
  year={2014},
  publisher={IEEE}
}

@article{ilpref,
  author={R. E. {N. Moraes} and C. C. {Ribeiro} and C. {Duhamel}},
  journal={IEEE Transactions on Wireless Communications}, 
  title={Optimal solutions for fault-tolerant topology control in wireless ad hoc networks}, 
  year={2009},
  volume={8},
  number={12},
  pages={5970-5981}}
  
@article{eswaran1976augmentation,
  title={Augmentation problems},
  author={Eswaran, Kapali P and Tarjan, R Endre},
  journal={SIAM Journal on Computing},
  volume={5},
  number={4},
  pages={653--665},
  year={1976},
  publisher={SIAM}
}

@article{central1,
  author={A. {Anand} and M. {Nithya} and T. {Sudarshan}},
  title={Coordination of mobile robots with master-slave architecture for a service application}, 
  journal={2014 International Conference on Contemporary Computing and Informatics (IC3I)}, 
  year={2014},
  volume={},
  number={},
  pages={539-543},
  doi={10.1109/IC3I.2014.7019647}
}
 
@article{central2,
  author={J. H. {Jung} and S. {Park} and S. {Kim}},
  title={Multi-robot path finding with wireless multihop communications}, 
  journal={IEEE Communications Magazine}, 
  year={2010},
  volume={48},
  number={7},
  pages={126-132},
  doi={10.1109/MCOM.2010.5496889}
}

@article{decentral1,
  author={M. {Zareh} and L. {Sabattini} and C. {Secchi}},
  journal={2016 IEEE 55th Conference on Decision and Control (CDC)}, 
  title={Decentralized biconnectivity conditions in multi-robot systems}, 
  year={2016},
  volume={},
  number={},
  pages={99-104},
  doi={10.1109/CDC.2016.7798253}
}


@article{decentral2,
author = {Sabattini, Lorenzo and Chopra, Nikhil and Secchi, Cristian},
title = {Decentralized Connectivity Maintenance for Cooperative Control of Mobile Robotic Systems},
journal = {Int. J. Rob. Res.},
year = {2013},
issue_date = {October 2013},
publisher = {Sage Publications, Inc.},
address = {USA},
volume = {32},
number = {12},
issn = {0278-3649},
url = {https://doi.org/10.1177/0278364913499085},
doi = {10.1177/0278364913499085},
pages = {1411–1423}
}

@software{grb,
    title={Gurobi Optimizer Reference Manual},
    author={Gurobi Optimization, LLC}, 
    year={2020},  
    url={http://www.gurobi.com}
}

@article{sabattini2013decentralized,
  title={Decentralized connectivity maintenance for cooperative control of mobile robotic systems},
  author={Sabattini, Lorenzo and Chopra, Nikhil and Secchi, Cristian},
  journal={The International Journal of Robotics Research},
  volume={32},
  number={12},
  pages={1411--1423},
  year={2013},
  publisher={SAGE Publications Sage UK: London, England}
}

@article{robuffo2013passivity,
  title={A passivity-based decentralized strategy for generalized connectivity maintenance},
  author={Robuffo Giordano, Paolo and Franchi, Antonio and Secchi, Cristian and B{\"u}lthoff, Heinrich H},
  journal={The International Journal of Robotics Research},
  volume={32},
  number={3},
  pages={299--323},
  year={2013},
  publisher={SAGE Publications Sage UK: London, England}
}

@article{zavlanos2011graph,
  title={Graph-theoretic connectivity control of mobile robot networks},
  author={Zavlanos, Michael M and Egerstedt, Magnus B and Pappas, George J},
  journal={Proceedings of the IEEE},
  volume={99},
  number={9},
  pages={1525--1540},
  year={2011},
  publisher={IEEE}
}

@article{varadharajan2020swarm,
  title={Swarm relays: Distributed self-healing ground-and-air connectivity chains},
  author={Varadharajan, Vivek Shankar and St-Onge, David and Adams, Bram and Beltrame, Giovanni},
  journal={IEEE Robotics and Automation Letters},
  volume={5},
  number={4},
  pages={5347--5354},
  year={2020},
  publisher={IEEE}
}

@article{hung2019hierarchical,
  title={Hierarchical distributed control for global network integrity preservation in multirobot systems},
  author={Hung, Pham Duy and Vinh, Tran Quang and Ngo, Trung Dung},
  journal={IEEE transactions on cybernetics},
  volume={50},
  number={3},
  pages={1278--1291},
  year={2019},
  publisher={IEEE}
}

@article{stephan2017concurrent,
  title={Concurrent control of mobility and communication in multirobot systems},
  author={Stephan, James and Fink, Jonathan and Kumar, Vijay and Ribeiro, Alejandro},
  journal={IEEE Transactions on Robotics},
  volume={33},
  number={5},
  pages={1248--1254},
  year={2017},
  publisher={IEEE}
}

@inproceedings{cornejo2010fault,
  title={Fault-tolerance through k-connectivity},
  author={Cornejo, Alejandro and Lynch, Nancy},
  booktitle={Workshop on network science and systems issues in multi-robot autonomy: ICRA},
  volume={2},
  pages={2010},
  year={2010},
  organization={Citeseer}
}

@inproceedings{luo2020minimally,
  title={Minimally disruptive connectivity enhancement for resilient multi-robot teams},
  author={Luo, Wenhao and Chakraborty, Nilanjan and Sycara, Katia},
  booktitle={2020 IEEE/RSJ International Conference on Intelligent Robots and Systems (IROS)},
  pages={11809--11816},
  year={2020},
  organization={IEEE}
}

@incollection{ghedini2018decentralized,
  title={A decentralized control strategy for resilient connectivity maintenance in multi-robot systems subject to failures},
  author={Ghedini, Cinara and Ribeiro, Carlos HC and Sabattini, Lorenzo},
  booktitle={Distributed Autonomous Robotic Systems},
  pages={89--102},
  year={2018},
  publisher={Springer}
}

@inproceedings{kuzminykh2017testing,
  title={Testing of communication range in ZigBee technology},
  author={Kuzminykh, Ievgeniia and Snihurov, Arkadii and Carlsson, Anders},
  booktitle={2017 14th International Conference The Experience of Designing and Application of CAD Systems in Microelectronics (CADSM)},
  pages={133--136},
  year={2017},
  organization={IEEE}
}

@inproceedings{iordache2017field,
  title={Field testing of Bluetooth and ZigBee technologies for vehicle-to-infrastructure applications},
  author={Iordache, Valentin and Gheorghiu, Razvan Andrei and Minea, Marius and Cormos, Angel Ciprian},
  booktitle={2017 13th International Conference on Advanced Technologies, Systems and Services in Telecommunications (TELSIKS)},
  pages={248--251},
  year={2017},
  organization={IEEE}
}

\begin{thebibliography}{10}
\providecommand{\url}[1]{#1}
\csname url@samestyle\endcsname
\providecommand{\newblock}{\relax}
\providecommand{\bibinfo}[2]{#2}
\providecommand{\BIBentrySTDinterwordspacing}{\spaceskip=0pt\relax}
\providecommand{\BIBentryALTinterwordstretchfactor}{4}
\providecommand{\BIBentryALTinterwordspacing}{\spaceskip=\fontdimen2\font plus
\BIBentryALTinterwordstretchfactor\fontdimen3\font minus
  \fontdimen4\font\relax}
\providecommand{\BIBforeignlanguage}[2]{{%
\expandafter\ifx\csname l@#1\endcsname\relax
\typeout{** WARNING: IEEEtran.bst: No hyphenation pattern has been}%
\typeout{** loaded for the language `#1'. Using the pattern for}%
\typeout{** the default language instead.}%
\else
\language=\csname l@#1\endcsname
\fi
#2}}
\providecommand{\BIBdecl}{\relax}
\BIBdecl

\bibitem{luo2019minimum}
W.~Luo and K.~Sycara, ``Minimum k-connectivity maintenance for robust
  multi-robot systems,'' in \emph{2019 IEEE/RSJ International Conference on
  Intelligent Robots and Systems (IROS)}.\hskip 1em plus 0.5em minus
  0.4em\relax IEEE, 2019, pp. 7370--7377.

\bibitem{atay2009mobile}
N.~Atay and B.~Bayazit, ``Mobile wireless sensor network connectivity repair
  with k-redundancy,'' in \emph{Algorithmic Foundation of Robotics VIII}.\hskip
  1em plus 0.5em minus 0.4em\relax Springer, 2009, pp. 35--49.

\bibitem{basu2004movement}
P.~Basu and J.~Redi, ``Movement control algorithms for realization of
  fault-tolerant ad hoc robot networks,'' \emph{IEEE network}, vol.~18, no.~4,
  pp. 36--44, 2004.

\bibitem{abbasi2008movement}
A.~A. Abbasi, M.~Younis, and K.~Akkaya, ``Movement-assisted connectivity
  restoration in wireless sensor and actor networks,'' \emph{IEEE Transactions
  on parallel and distributed systems}, vol.~20, no.~9, pp. 1366--1379, 2008.

\bibitem{liu2010simple}
H.~Liu, X.~Chu, Y.-W. Leung, and R.~Du, ``Simple movement control algorithm for
  bi-connectivity in robotic sensor networks,'' \emph{IEEE Journal on Selected
  Areas in Communications}, vol.~28, no.~7, pp. 994--1005, 2010.

\bibitem{lee2015connectivity}
S.~Lee, M.~Younis, and M.~Lee, ``Connectivity restoration in a partitioned
  wireless sensor network with assured fault tolerance,'' \emph{Ad Hoc
  Networks}, vol.~24, pp. 1--19, 2015.

\bibitem{ilpref}
R.~E. {N. Moraes}, C.~C. {Ribeiro}, and C.~{Duhamel}, ``Optimal solutions for
  fault-tolerant topology control in wireless ad hoc networks,'' \emph{IEEE
  Transactions on Wireless Communications}, vol.~8, no.~12, pp. 5970--5981,
  2009.

\bibitem{eswaran1976augmentation}
K.~P. Eswaran and R.~E. Tarjan, ``Augmentation problems,'' \emph{SIAM Journal
  on Computing}, vol.~5, no.~4, pp. 653--665, 1976.

\bibitem{frederickson1981approximation}
G.~N. Frederickson and J.~Ja’Ja’, ``Approximation algorithms for several
  graph augmentation problems,'' \emph{SIAM Journal on Computing}, vol.~10,
  no.~2, pp. 270--283, 1981.

\bibitem{khuller1993approximation}
S.~Khuller and R.~Thurimella, ``Approximation algorithms for graph
  augmentation,'' \emph{Journal of algorithms}, vol.~14, no.~2, pp. 214--225,
  1993.

\bibitem{sabattini2013decentralized}
L.~Sabattini, N.~Chopra, and C.~Secchi, ``Decentralized connectivity
  maintenance for cooperative control of mobile robotic systems,'' \emph{The
  International Journal of Robotics Research}, vol.~32, no.~12, pp. 1411--1423,
  2013.

\bibitem{robuffo2013passivity}
P.~Robuffo~Giordano, A.~Franchi, C.~Secchi, and H.~H. B{\"u}lthoff, ``A
  passivity-based decentralized strategy for generalized connectivity
  maintenance,'' \emph{The International Journal of Robotics Research},
  vol.~32, no.~3, pp. 299--323, 2013.

\bibitem{zavlanos2011graph}
M.~M. Zavlanos, M.~B. Egerstedt, and G.~J. Pappas, ``Graph-theoretic
  connectivity control of mobile robot networks,'' \emph{Proceedings of the
  IEEE}, vol.~99, no.~9, pp. 1525--1540, 2011.

\bibitem{minelli2020self}
M.~Minelli, J.~Panerati, M.~Kaufmann, C.~Ghedini, G.~Beltrame, and
  L.~Sabattini, ``Self-optimization of resilient topologies for fallible
  multi-robots,'' \emph{Robotics and Auton. Systems}, vol. 124, p. 103384,
  2020.

\bibitem{panerati2019robust}
J.~Panerati, M.~Minelli, C.~Ghedini, L.~Meyer, M.~Kaufmann, L.~Sabattini, and
  G.~Beltrame, ``Robust connectivity maintenance for fallible robots,''
  \emph{Autonomous Robots}, vol.~43, no.~3, pp. 769--787, 2019.

\bibitem{cornejo2010fault}
A.~Cornejo and N.~Lynch, ``Fault-tolerance through k-connectivity,'' in
  \emph{Workshop on network science and systems issues in multi-robot autonomy:
  ICRA}, vol.~2.\hskip 1em plus 0.5em minus 0.4em\relax Citeseer, 2010, p.
  2010.

\bibitem{varadharajan2020swarm}
V.~S. Varadharajan, D.~St-Onge, B.~Adams, and G.~Beltrame, ``Swarm relays:
  Distributed self-healing ground-and-air connectivity chains,'' \emph{IEEE
  Robotics and Automation Letters}, vol.~5, no.~4, pp. 5347--5354, 2020.

\bibitem{hung2019hierarchical}
P.~D. Hung, T.~Q. Vinh, and T.~D. Ngo, ``Hierarchical distributed control for
  global network integrity preservation in multirobot systems,'' \emph{IEEE
  transactions on cybernetics}, vol.~50, no.~3, pp. 1278--1291, 2019.

\bibitem{stephan2017concurrent}
J.~Stephan, J.~Fink, V.~Kumar, and A.~Ribeiro, ``Concurrent control of mobility
  and communication in multirobot systems,'' \emph{IEEE Transactions on
  Robotics}, vol.~33, no.~5, pp. 1248--1254, 2017.

\bibitem{wang2010movement}
S.~Wang, X.~Mao, S.-J. Tang, X.~Li, J.~Zhao, and G.~Dai, ``On
  “movement-assisted connectivity restoration in wireless sensor and actor
  networks”,'' \emph{IEEE Transactions on Parallel and Distributed Systems},
  vol.~22, no.~4, pp. 687--694, 2010.

\bibitem{luo2020minimally}
W.~Luo, N.~Chakraborty, and K.~Sycara, ``Minimally disruptive connectivity
  enhancement for resilient multi-robot teams,'' in \emph{2020 IEEE/RSJ
  International Conference on Intelligent Robots and Systems (IROS)}.\hskip 1em
  plus 0.5em minus 0.4em\relax IEEE, 2020, pp. 11\,809--11\,816.

\bibitem{ghedini2018decentralized}
C.~Ghedini, C.~H. Ribeiro, and L.~Sabattini, ``A decentralized control strategy
  for resilient connectivity maintenance in multi-robot systems subject to
  failures,'' in \emph{Distributed Autonomous Robotic Systems}.\hskip 1em plus
  0.5em minus 0.4em\relax Springer, 2018, pp. 89--102.

\bibitem{kuzminykh2017testing}
I.~Kuzminykh, A.~Snihurov, and A.~Carlsson, ``Testing of communication range in
  zigbee technology,'' in \emph{2017 14th International Conference The
  Experience of Designing and Application of CAD Systems in Microelectronics
  (CADSM)}.\hskip 1em plus 0.5em minus 0.4em\relax IEEE, 2017, pp. 133--136.

\bibitem{iordache2017field}
V.~Iordache, R.~A. Gheorghiu, M.~Minea, and A.~C. Cormos, ``Field testing of
  bluetooth and zigbee technologies for vehicle-to-infrastructure
  applications,'' in \emph{2017 13th International Conference on Advanced
  Technologies, Systems and Services in Telecommunications (TELSIKS)}.\hskip
  1em plus 0.5em minus 0.4em\relax IEEE, 2017, pp. 248--251.

\bibitem{west2001introduction}
D.~B. West \emph{et~al.}, \emph{Introduction to graph theory}.\hskip 1em plus
  0.5em minus 0.4em\relax Prentice hall Upper Saddle River, 2001, vol.~2.

\bibitem{grb}
\BIBentryALTinterwordspacing
L.~Gurobi~Optimization, ``Gurobi optimizer reference manual,'' 2020. [Online].
  Available: \url{http://www.gurobi.com}
\BIBentrySTDinterwordspacing

\bibitem{rabban2021failure}
\BIBentryALTinterwordspacing
M.~Ishat-E-Rabban, P.~Tokekar, ``Failure-resilient coverage maximization with multiple robots'', \emph{IEEE Robotics and Automation Letters}, vol.~6(2), pp.~3894-3901, 2021.

\bibitem{rabban2019mvfs}
\BIBentryALTinterwordspacing
M.~Rabban, M.~Ali, M.~Cheema, T.~Hashem, ``The Maximum Visibility Facility Selection Query in Spatial Databases'', \emph{27th ACM Sigspatial International Conference on Advances in Geographic Information Systems}, pp.~149--158, 2019.


\bibitem{shi2021communication}
\BIBentryALTinterwordspacing
G.~Shi, I.E.~Rabban, L.~Zhou, P.~Tokekar, ``Communication-aware multi-robot coordination with submodular maximization'', \emph{2021 IEEE International Conference on Robotics and Automation (ICRA)}, pp.~8955-8961). 2021.
\BIBentrySTDinterwordspacing

\end{thebibliography}





\end{document}